\documentclass[inte,nonblindrev]{informs3}

\OneAndAHalfSpacedXII

\usepackage{hyperref}
\usepackage{booktabs}

\newcommand{\pol}{\href{https://www.reddit.com/r/politics/}{/r/politics }}
\newcommand{\wn}{\href{https://www.reddit.com/r/worldnews/}{/r/worldnews }}
\newcommand{\td}{\href{https://www.reddit.com/r/The\_Donald/}{/r/The\_Donald }}
\newcommand{\rdt}{\href{https://www.reddit.com/}{Reddit }}
\newcommand{\subr}[1]{\href{https://www.reddit.com/r/#1}{#1 }}

\usepackage[english]{babel}

\usepackage{lipsum}

\usepackage{changepage}

\usepackage{graphicx}
\usepackage{float}

\usepackage{rotating}
\usepackage{tikz}
\usetikzlibrary{
  arrows,
  backgrounds,
  cd,
  shadows,
  shapes,
  topaths,
  decorations.pathreplacing
}
\tikzset{
    position label/.style={
       below = 3pt,
       text height = 1.5ex,
       text depth = 1ex
    },
   brace/.style={
     decoration={brace, mirror},
     decorate
   }
}
\usepackage{comment}

\usepackage[colorinlistoftodos,shadow,backgroundcolor=orange,linecolor=red]{todonotes}

\usepackage{sansmathfonts}
\usepackage{nicefrac}
\usepackage{units}
\usepackage{amsmath, amssymb, amsfonts}
\usepackage{mathtools}
\usepackage{oubraces}

\usepackage{cleveref}

\usepackage{natbib}
 \bibpunct[, ]{(}{)}{,}{a}{}{,}%
 \def\bibfont{\small}%
 \def\newblock{\ }%

\TheoremsNumberedThrough   
\EquationsNumberedThrough

\begin{document}

\TITLE{Creolizing the Web}

\ARTICLEAUTHORS{
\AUTHOR{Abhinav Tamaskar}
\AFF{Courant Institute of Mathematical Sciences, NYU, \EMAIL{tamaskar@cs.nyu.edu}}
\AUTHOR{Roy Rinberg}
\AFF{Courant Institute of Mathematical Sciences, NYU, \EMAIL{royrinberg@gmail.com}}
\AUTHOR{Sunandan Chakraborty}
\AFF{Indiana University, NYU, \EMAIL{sunandanchakraborty@gmail.com}}
\AUTHOR{Bud Mishra}
\AFF{Courant Institute of Mathematical Sciences, NYU, \EMAIL{mishra@cs.nyu.edu}}
}
\ABSTRACT{
The evolution of language has been a hotly debated subject with contradicting hypotheses and unreliable claims. Drawing from signalling games, dynamic population mechanics, machine learning and algebraic topology, we present a method for detecting evolutionary patterns in a sociological model of language evolution. We develop a minimalistic model that provides a rigorous base for any generalized evolutionary model for language based on communication between individuals. We also discuss theoretical guarantees of this model, ranging from stability of language representations to fast convergence of language by temporal communication and language drift in an interactive setting. Further we present empirical results and their interpretations on a real world dataset from \rdt to identify communities and echo chambers for opinions, thus placing obstructions to reliable communication among communities.
}%

\maketitle

\section*{Introduction}

The mystery of language evolution and its (co-)evolution with learning continues to arouse intense debates. There are only a handful of conceptual frameworks for human languages that have found common acceptance: (i) Human language is a biological artefact, as opposed to a cultural artifact \cite{lenneberg1967biological}. (ii) Human language builds on a hierarchical structure, whose depth is not upper-bounded\cite{christiansen1999toward}. (iii) Human language acquisition occurs over a surprisingly short period aided primarily by positive examples \cite{briscoe2002linguistic}\cite{li1996li}. However, there are many other corollaries that seem to have found neither acceptance in theory nor utilization in tool-boxes that aim to automate natural language processing.

There are other similar questions in the biology of evolution: e.g., codon evolution and evolution of intercellular signaling, which are important in the emergence of cellularization and multi-cellular organisms, respectively \cite{sharp1994codon}. The theoretical framework for them can be built on information-asymmetric games and their conventional Nash equilibria, and can be tested experimentally in artificial cells with unnatural bases (and the resulting codons), and in modified cells with chimeric receptors, for instance. There are few natural experiments that shed light on these processes, e.g., mitochondria and tumor cells, and they have also played an important role in our understanding of evolution of these systems\cite{acevedo2014mutational}\cite{korolev2014turning}.

These systems, like human language, can also be thought of encoding some form of inter-agent coordination (not necessarily faithfully)\cite{traulsen2009exploration}. They also share few other traits: e.g., (i) Universality, (ii) Stability and (iii) Near Optimality (with respect to suitably selected utility); we will call them USNO-theories. A rigorous theory for human languages may seek to build on similar traits: (i) A universal grammar (with some flexibility for parametrization)\cite{cook2014chomsky}, (ii) Stability (with faithful acquisition using meager amount of positive stimuli)\cite{eisner2006perceptual}\cite{nichols1992linguistic} and (iii) Near Optimality (as a solution to minimal design specifications)\cite{escudero2005linguistic}\cite{case2008optimal}. However, hypotheses related to physiology of a language organ or the genetics of linguistic phenotypes are not readily testable experimentally as human language is unique to humans thus imposing stringent ethical barriers against their experimental manipulation. Some analysis of bird-songs have been useful, but not very conclusive (for obvious reasons). In silico models that work reasonably in the context of machine learning and artificial intelligence have focused on large text corpora and semi-supervised learning (with massive number of counter-examples) that do not capture the human context and remain orthogonal to the biology of languages\cite{collobert2008unified}.

\begin{figure}[ht]
\centering
\includegraphics[width=\textwidth]{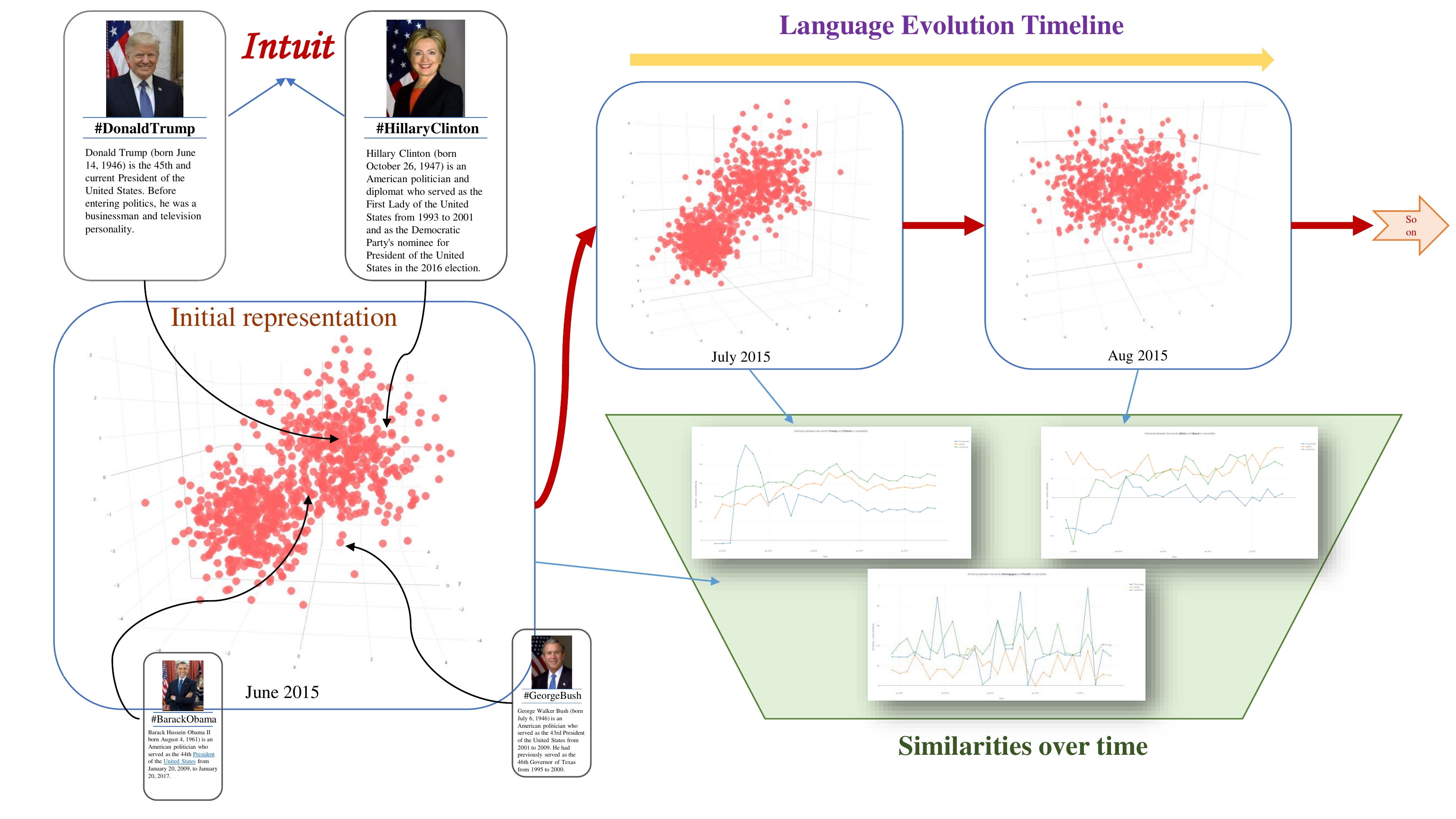}
\caption{\label{fig:intuit} A linguistic object {\tt intuit}  consists of an image, a hashtag and a short description of the intuit. A language starts with a small number of such intuits in a core germinal population and accrues additional users who add additional intuits, and use them to communicate. External to this dynamics, we can examine the time-stamped representations of the vocabulary of intuits and observe the evolution of the representation through time. This analysis will help us in understanding the social vs.~inherent evolution of the representations and of language, based on the changes of the similarities of the intuits over time. Analyzing the data on a per user basis will give us hitherto unknown knowledge of the socio-cultural effects of interactions and community effects on dialects of the language. But, motivated to study the language as a whole, as opposed to just a pair of intuits and their similarities, we were led to novel mathematics to analyze topological differences in representations over time.}
\end{figure}

Interesting natural experiments that are thought to have lent support to USNO-theories are in the creolization process, where a group of individuals from Old World are assembled with no common human language to use for coordination, but who give rise to a second generation of New World speech community that invent a human language (Creole) with a new parametrization of the universal grammar, but also enjoying the stability and near-optimality that is common to already-existing human languages\cite{siegel1997mixing}. However, while Creole languages can be studied, their evolution remains poorly understood as there exists no data recording their historical dynamics\cite{hymes1971pidginization}. As Crick's Frozen Accident hypothesis and the Cambrian explosion have been used to explain codon evolution or multi-cellularity, there has been human language evolution's Pop hypothesis that suggests creolization would happen suddenly and freeze quickly, not thawing ever again\cite{koonin2009origin}. The alternative experimentally-supported hypothesis suggesting emergence of a human language as a stable separating Nash equilibria of an information asymmetric game would be more explanatory and hence appealing\cite{huang2001spoken}\cite{chomsky1972stages}.

Motivated thus, we have proposed using crowd-sourcing to create a super sized speech-community with a massively scalable socio-technological version of creolization. The elements of these systems would be \textbf{intuits} (with more details in later sections), and eventually a grammar that linearizes (or even planarizes) Intuits in a stable manner. We call this idea ``Creolization of the Web'' and here, we study various algorithmic issues related to machine learning, natural language processing and evolutionary processes to study the feasibility of such a creolization experiment(s). In particular, we focus on (i) definition of Intuits, the building blocks of the creolization combining images, hashtags, and short tweet-like (140 characters) description, (ii) their dynamic geometric representation and (iii) evolution of the representation via a Bayesian echo chamber. We illustrate the process with \rdt data involving political subreddits to identify evolutionary patterns that emerge in a dynamic population interaction model (\cref{fig:intuit}, \cite{guo2015bayesian}).

We realize that the ultimate system that combines elements of wiki, Twitter, emoticons, and Facebook could provide enormous utility in web-search, social networking, and shared economy, possibly displacing English as the defacto intermediate language of the web. Creation of a suitable infrastructure for Intuits remains a secondary but critical goal.
\section*{Problem Description}

We aim to build a database of a pictorial language called, \textbf{intuits}, which will help in the process of learning language evolution. The building blocks of this language, called an \emph{intuit}, is a token for any word in the vocabulary, where the token contains richer information than just the word, by storing (1) a title (a hashtag unique identifier, (2) a brief (140 character) description of the title and (3) an image of the title. The presence of this database to track the change of meanings of the intuits over time will give important insights to the theory of language evolution.

\begin{figure}[ht]
    \centering
    \includegraphics[width=0.8\textwidth]{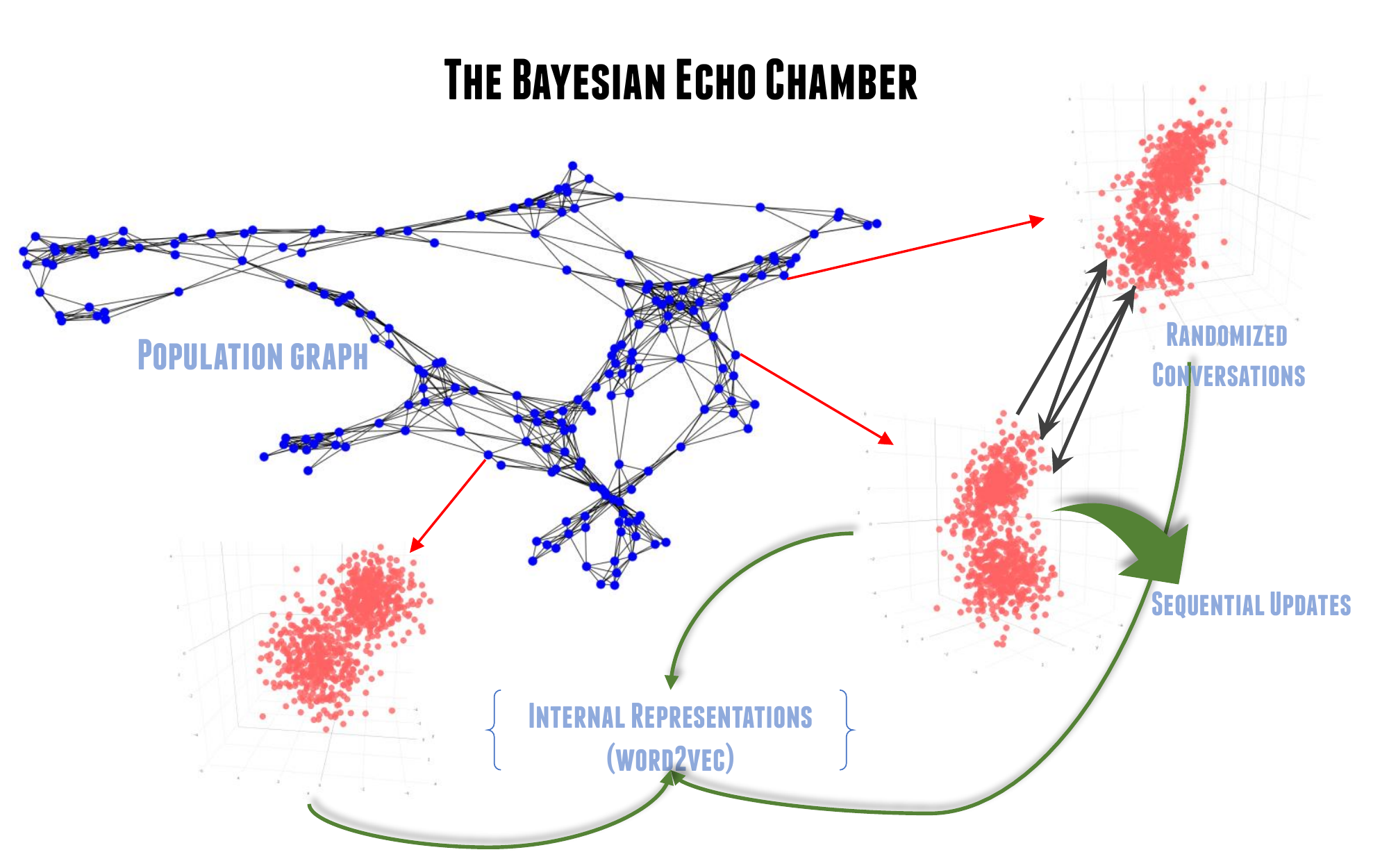}
    \caption{The Bayesian Echo Chamber. In the model of the Bayesian Echo Chamber, the population is represented as a graph of individuals, called (language) learners, where the edges denote interactions (conversations) between learners. Each learner has their own internal representation of the language, which they use in their conversations with their neighbors. The conversations happen based on a particular topic. And the words in the conversation are chosen based on the similarities of the words with the topic in the internal representation of the learner.\label{fig:bec}}
\end{figure}

In this paper we give a baseline minimal model, based on the \emph{Bayesian Echo Chamber}\cite{guo2015bayesian}, which is applicable to any evolutionary method and also has the flexibility to be individualized to any language using concrete grammars and objective semantics specific to that language. To experimentally verify the plausibility of such a model, we analyze real world data from \rdt, which is an online community of users -- sufficiently active and engaged to model  communication interactions in a population. \rdt is structured as a collection of ``subreddits'', which are communities dedicated to a particular topic, such as \subr{gaming}, \subr{sports}, \subr{technology}, etc. Each user of \rdt is generally subscribed to a few of the subreddits, focusing on the content that the user generally browses and is exposed to. The \rdt community has been frequently divided on many topics, most recent of which has been on the political spectrum. This discordance provides a very rich environment to measure the effects of social stratification of language due to dissenting views between communities. We started with a synthetic model of intuits for a large population interacting panmictically (or as determined by an expander population graph) as it provides a baseline for an idealized theoretical model and a null model for hypotheses testing.

\begin{figure}[ht]
    \centering
    \includegraphics[width=0.45\textwidth]{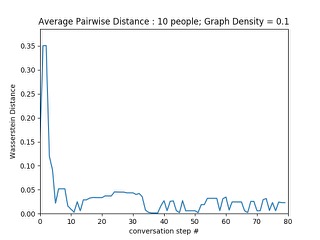}
    \includegraphics[width=0.45\textwidth]{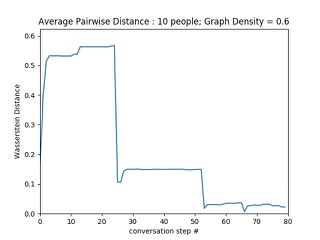}\\
    \includegraphics[width=0.45\textwidth]{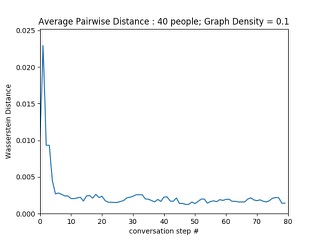}
    \includegraphics[width=0.45\textwidth]{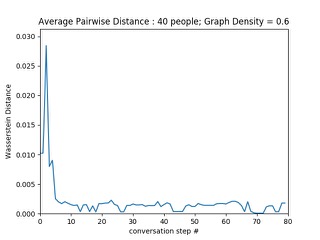}
    \caption{Simulation of a population with various parameters of connectivity and size. The simulations show a fast convergence in the representation of individuals and a small drift over time after convergence. These results agree with the accepted theories of language evolution which predict fast stabilization and small drifts in language representations \cite{greenberg1960quantitative}\cite{fischer1958social}.}
    \label{fig:bec2}
\end{figure}

The change in language is measured using computational tools (originally developed for Natural Language Processing, NLP), specifically {\tt word2vec}, to get a feature rich, high dimensional embedding of the elements of a language associated with individual speakers. These embeddings can be thought of as the representation of the language for the individual and the difference in the representations gives us a measure of the dissimilarity between the interpretations of the language in the population. Each representation being a corpus of high dimensional points (``point clouds''), there is no standard notion of a distance between two such comparable representations. We propose to apply a topological metric using \emph{persistent homology}\cite{edelsbrunner2008persistent}\cite{carlsson2009topology}, which is an emerging field of computational mathematics, quantifying a sense of difference between two representations. The advantages of using the topological metric is the rich information content, which provides insights into the local features of a space as well as measuring the global differences between two representations\cite{arora2006local}\cite{singh2007topological}.

\begin{figure}[ht]
    \centering
    \includegraphics[width=\textwidth]{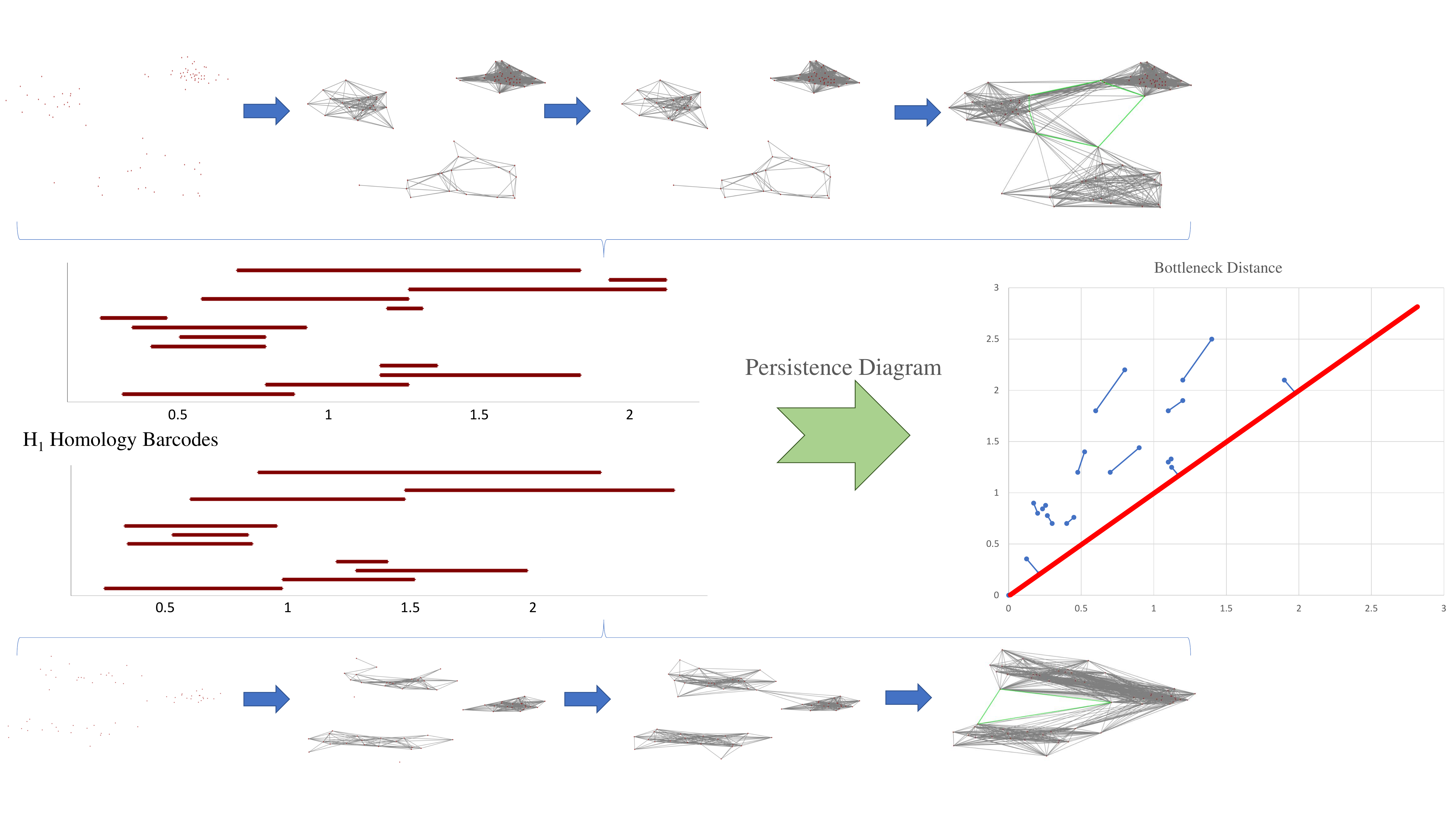}
    \caption{\label{fig:tdaoutline}Topological distance of embeddings. To define the distance between representations we start with a topological metric, as it gives information about the representation as well as the differences between two representations. We examine features in the word2vec embeddings of the vocabulary of a learner and calculate the distances based on the geometric embeddings. Two words will be close to each other in the word2vec space iff they are semantically nearly synonymous in the vocabulary of the learner. Now we can calculate the persistent homologies of the embedding and obtain the persistent diagrams of the space. This computation gives us the \textbf{bottleneck distance} between the two diagrams, and equips us with a sense of how dis-similar two embeddings are.}
\end{figure}

\section*{Results}

\subsection*{Confirmation of Echo Chambers in \rdt}

\begin{figure}[ht]
\centering
\includegraphics[width=0.98\textwidth]{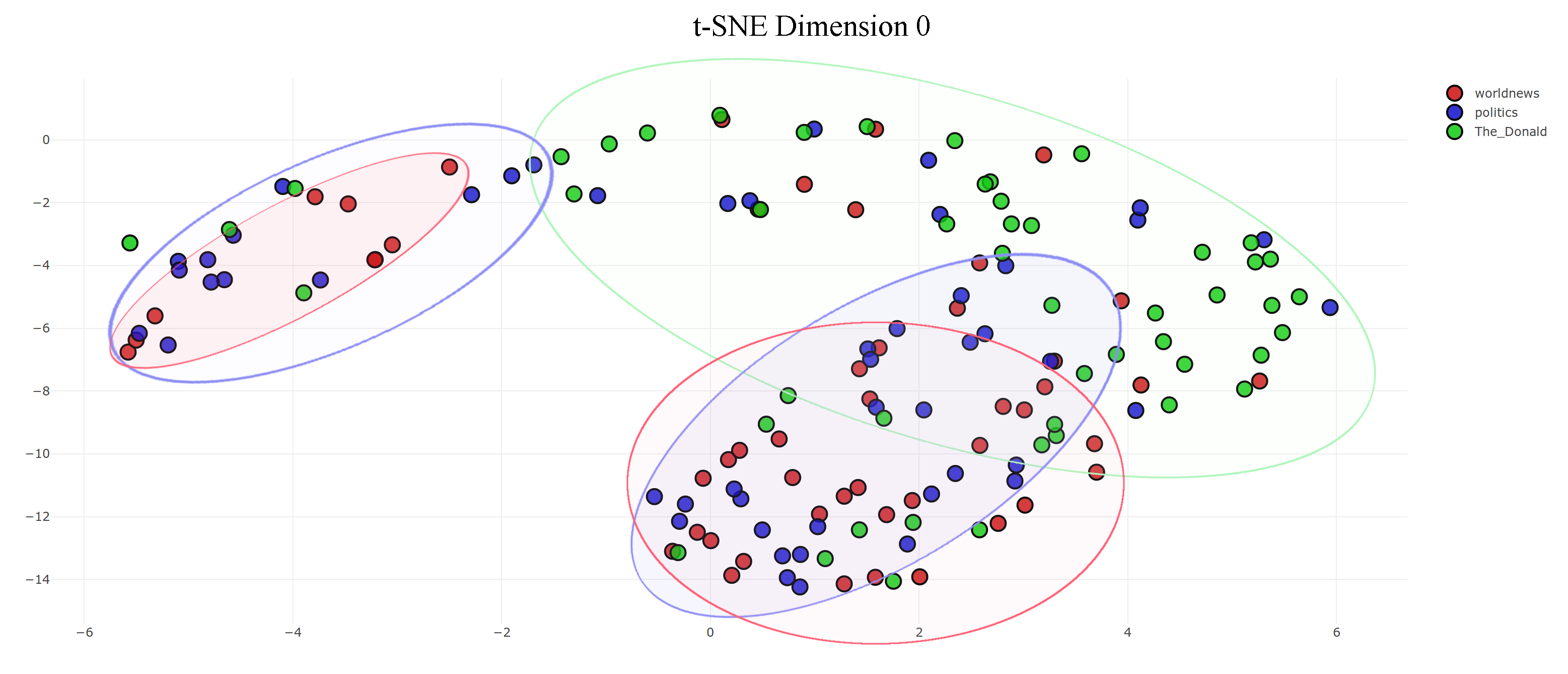}\\
\includegraphics[width=0.48\textwidth]{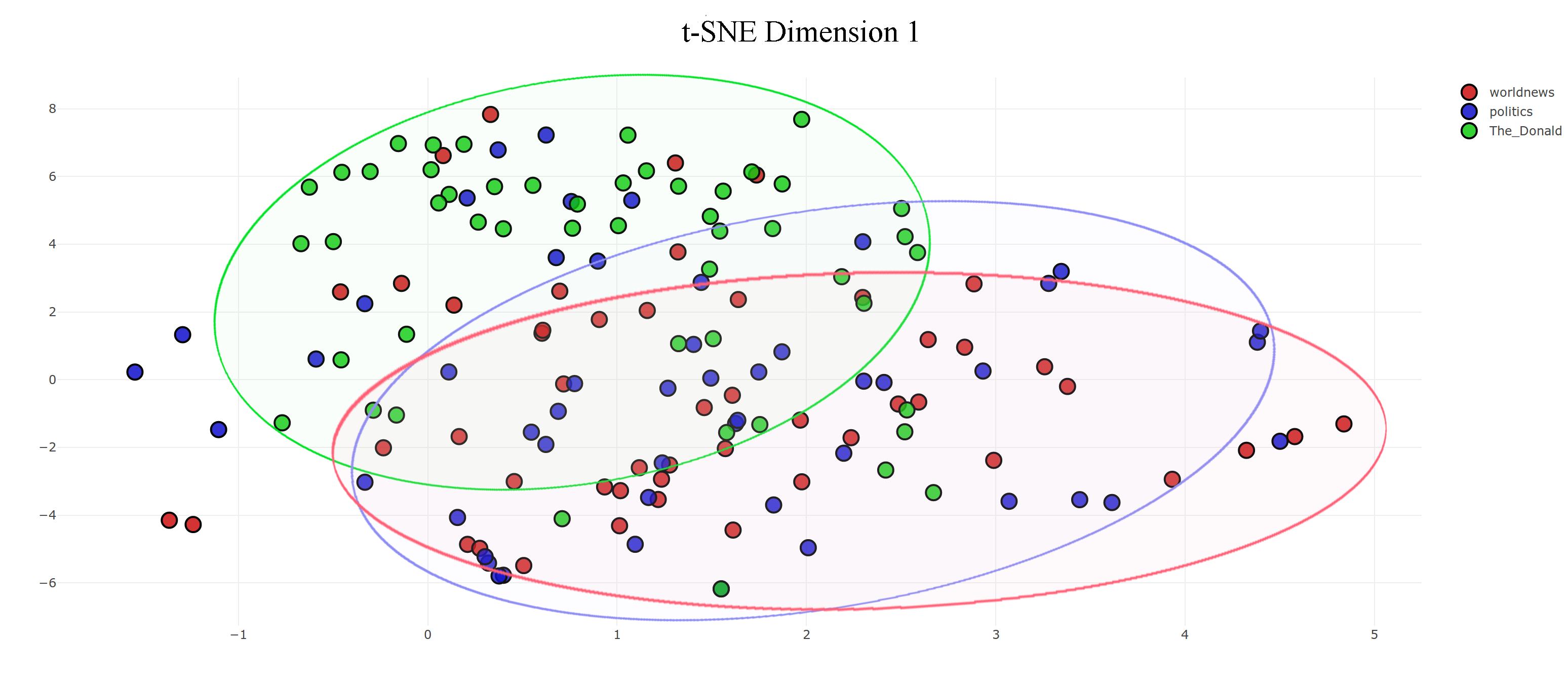}
\includegraphics[width=0.48\textwidth]{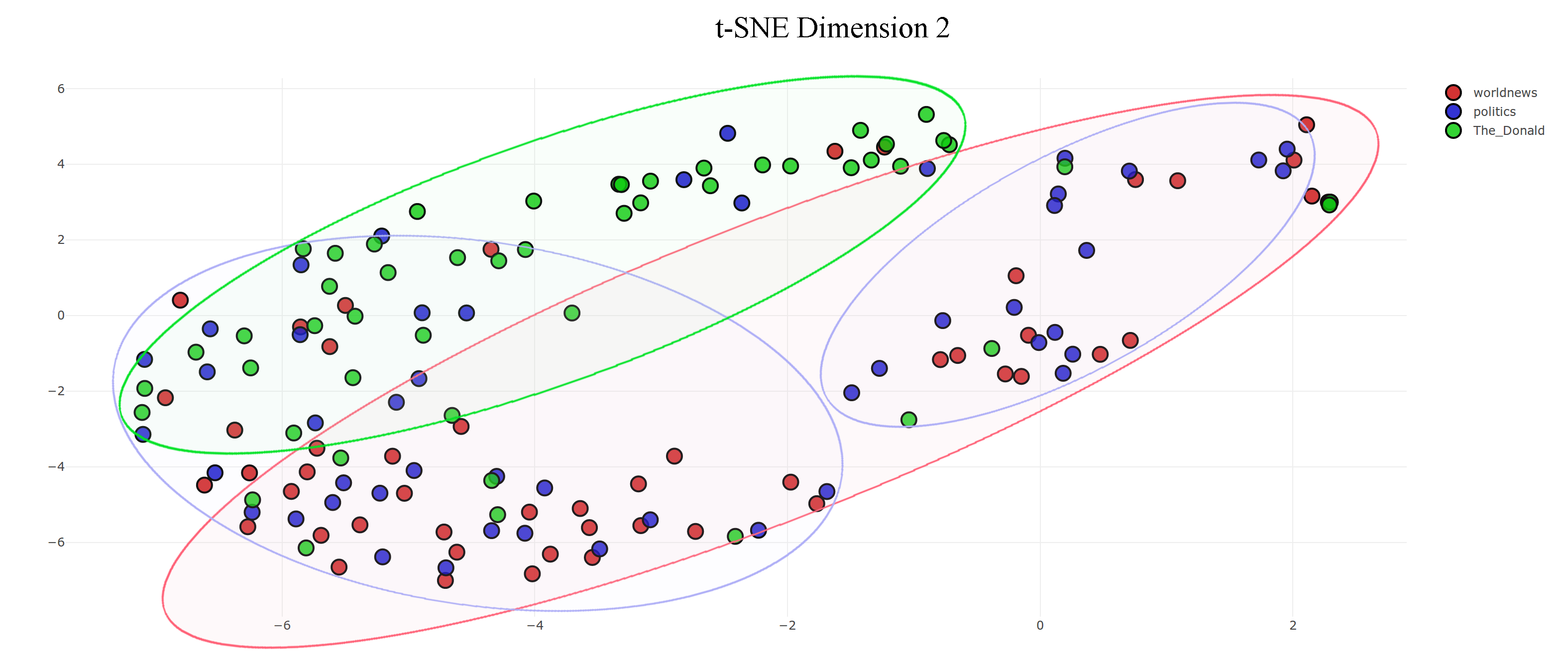}
\caption{t-SNE of users in different subreddits\label{fig:embed}. The embeddings of each user were generated using the state of the art {\tt word2vec} models and their \rdt data from the June 2015 till November 2017; using which, we calculated the distances between each pair of users using the persistent homology metric. To visualize and quantify the clusters formed using this metric, we performed t-SNE in 2-D plane, as t-SNE gives higher probabilities to cluster pairs which have small distance while not clustering larger distance pairs. The resulting clusters show a stark similarity between the users of \pol and \wn, while those of \td are clustered separately. This behavior is mimicked in all dimensions, showing that there is little communication happening between these communities.}
\end{figure}

The existence of echo chambers in any society can be manifested in many forms, such as the presence of dialects across the physical distribution of a population or the prevalence of accepted norms and
ideologies in a community. The frequent divide in the political spectrum within a population, popularly described as the ``left and right-wing extremisms'', is an interesting part of language that can be harnessed to understand political ideologies in subreddits.

To examine this hypothesis explaining a spectrum in the communities, we proceeded to analyze the three most popular political subreddits which are widely believed to cater to different groups, namely \pol, \wn, \td. \pol is the subreddit focused on US politics; the user base of \pol has been thought to be largely liberal. \wn focuses more on international news and has frequent discussions on international relations between countries. \td is another US politics focused group, which was founded in June 2015, and has a more republican user base.

We collected the top fifty most frequent and popular users from each subreddit, to infer a model of the user base of the subreddit. We took the \rdt data for each user over a period of two years from June 2015 to November 2017. Using this as a data corpus for the {\tt word2vec} model we created word embeddings for each user to get a point cloud of the vocabulary of the user. Persistent homology was then used to calculate the \emph{barcodes} of the {\tt word2vec} embeddings of each user. Based on the barcodes of each user, the bottleneck distance metric provided a similarity score to every pair of users, which was used by t-SNE to get a low-dimensional clustering embedding of the population \cref{fig:embed}. The advantage of the t-SNE clustering is the ability to find highly probable clusters (i.e., with a large likelihood), while low probability clusters are ignored.

Based on the t-SNE clusterings, we see a stark similarity between the users of \pol and \wn. This structure not only supports the hypothesis postulating existence of largely liberal user bases in the two subreddits, but also gives a clear method to find echo chambers across the whole \rdt community. The users of \td are shown to be hugely dissimilar to those of \pol and \wn as the political ideologies of republicans have many contrasting accepted notions than those of democrats.

The idea for using these embeddings and the topological similarity can also be extended to any other spatial model, such as the embeddings computed by {\tt GloVE}, {\tt fasttext}, {\tt sense}, etc. {\tt Sense} embeddings have the additional characteristic of being able to identify polysemy. Thus Topological Data Analysis (TDA) can take advantage of this feature to characterize measures of polysemy between different languages. Nonetheless, one needs to be careful, when considering the potential effects of prevalent topics in the subreddits and to ensure that secondary structures do not dominate the embedding criterion. This goal can be ensured by restricting the topic base to a particular subset so that the vocabulary of the topics remains largely consistent through the subreddits.

\subsection*{Comparison of subreddits gives details of divergence over time}

\begin{figure}[ht]
    \centering
    \includegraphics[width=0.96\textwidth]{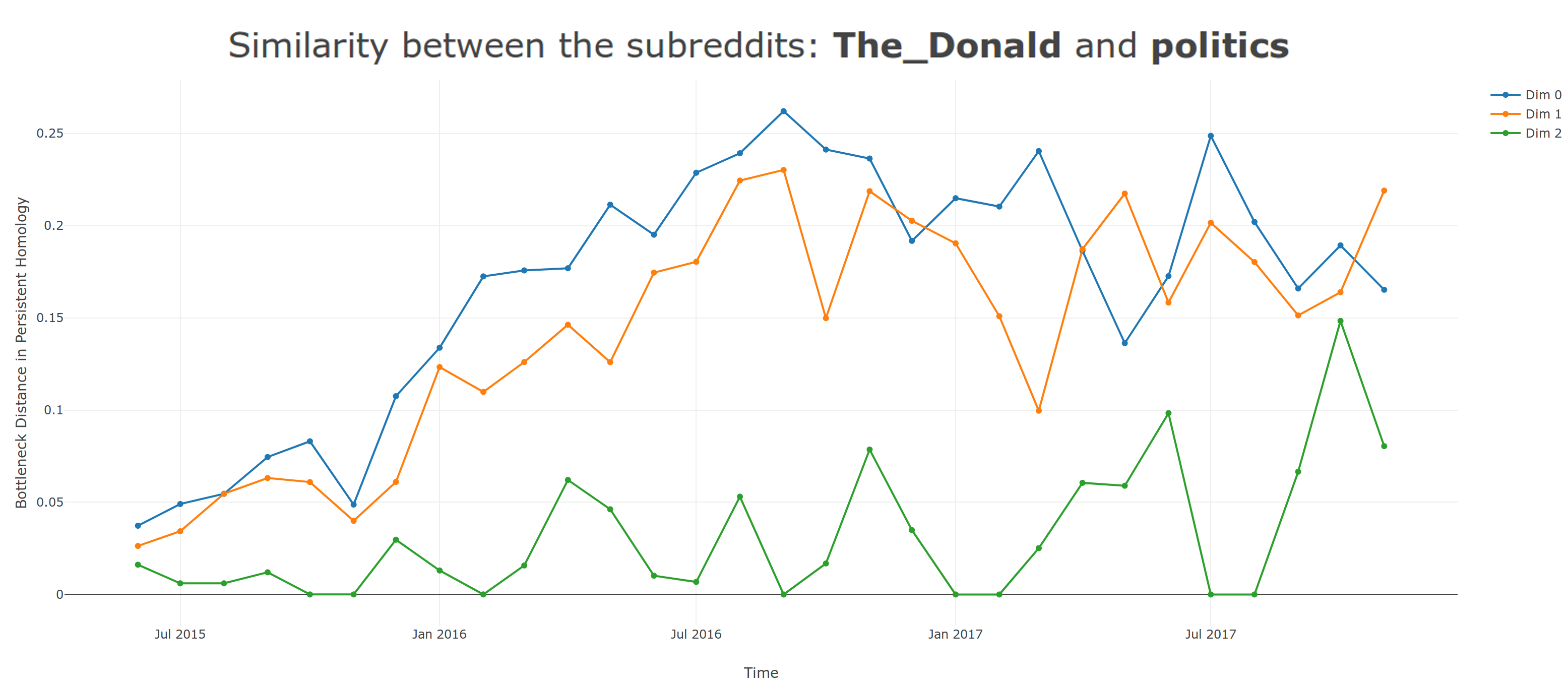}
    \includegraphics[width=0.48\textwidth]{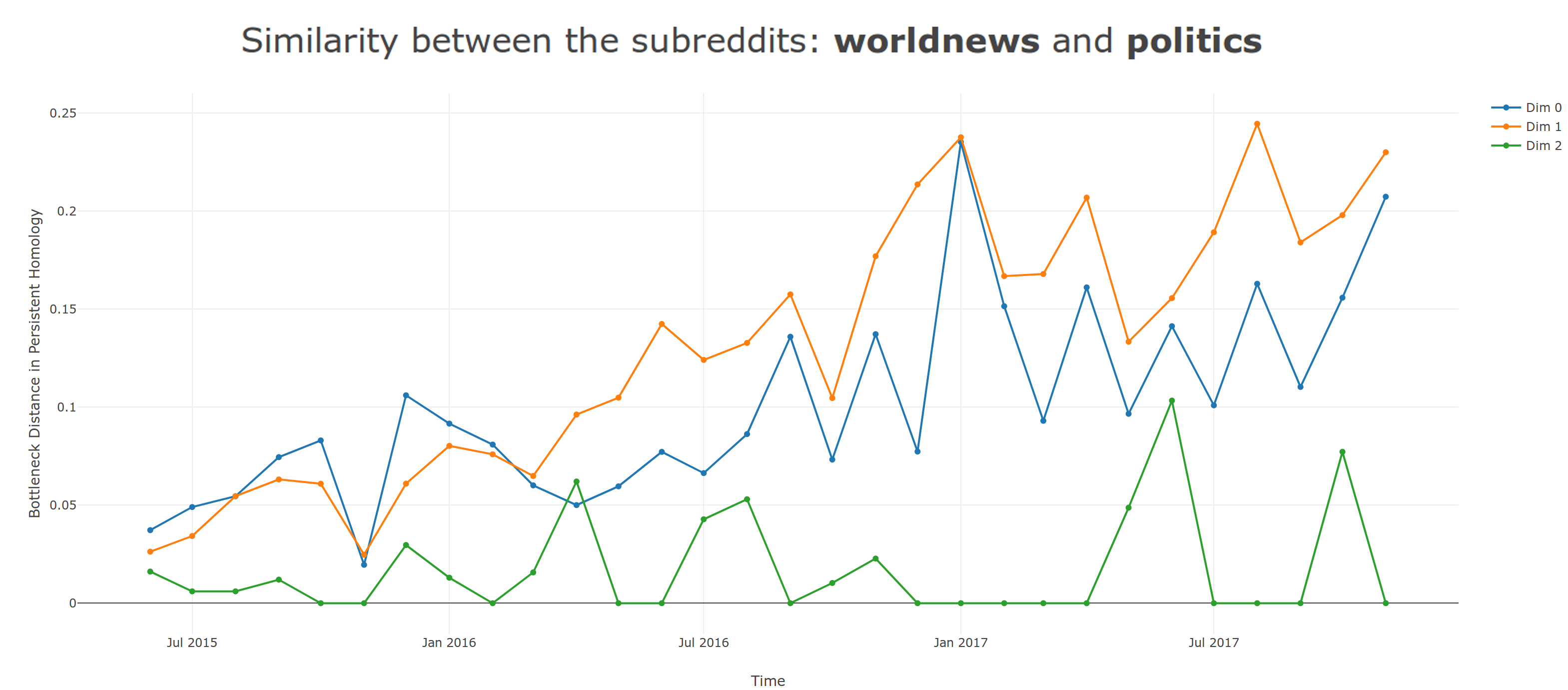}
    \includegraphics[width=0.48\textwidth]{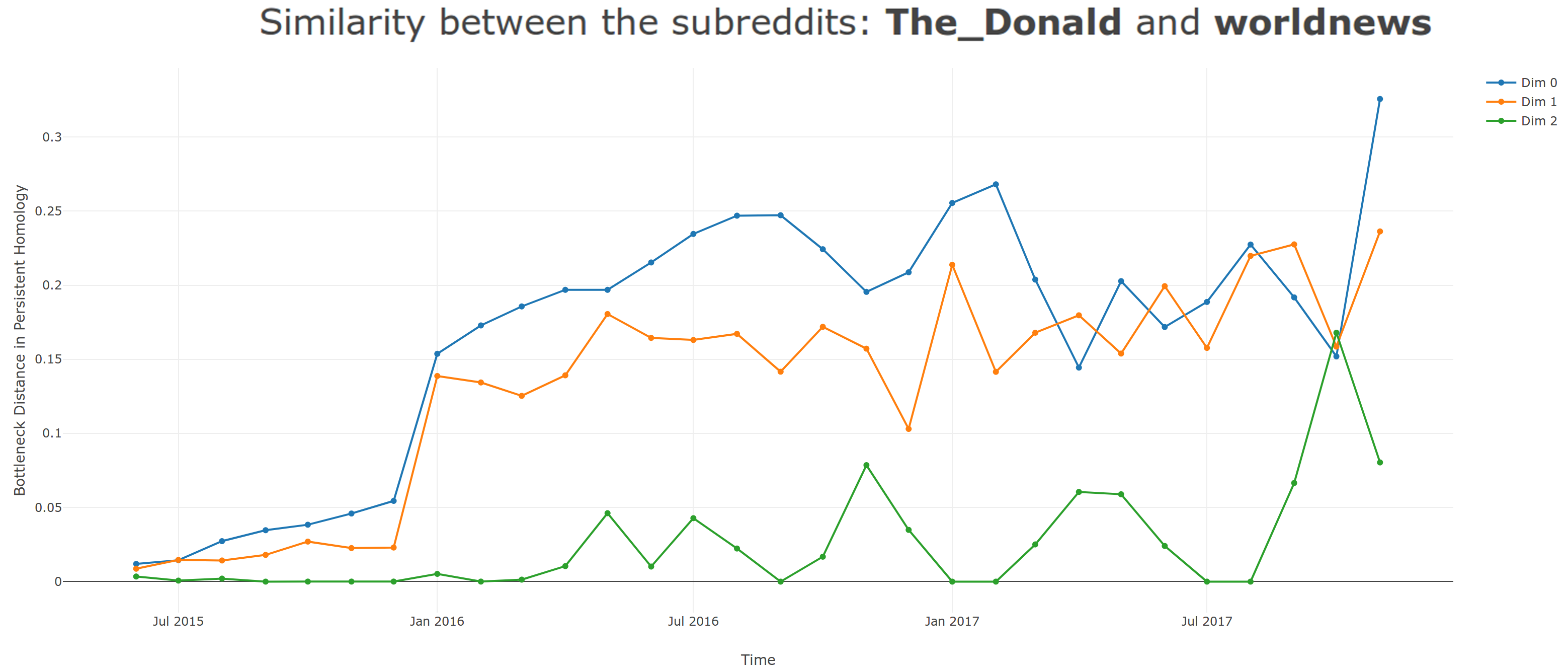}
    \caption{Bottleneck distance between subreddits\label{fig:subsdist}. We collect the most popular posts from every month in each subreddit to build a temporal model for language representation. Using the bottleneck distance of persistence diagrams we can calculate the distance between the language representation over time and see the effects of the community structure. The consistent increase in the distance between the representations confirms the hypothesis of echo chambers in subreddits, leading to a divergence in representations and topic focus between the subreddits.}
\end{figure}

One of the main reasons for performing temporal analysis of language in \rdt is to be able to identify the effects of communications (or lack thereof) between the population on the language of each community. To analyze this effect, we took the most popular topics from each month, from June 2015 till November 2017, in each subreddit and made an incremental {\tt word2vec} model. This incremental model presented to us a highly dynamic picture of each subreddit through time, which we used as an input to the persistent homology toolbox to rigorously quantify the changing similarities over time \cref{fig:subsdist}.

We observe that there is a consistent increase in the relative pair-wise distances of the subreddits. This dispersion corresponds to the formation of communities and how the nascent communities differ in interpreting semantic nature and sentiments of words in the subreddits. The increase in the bottleneck distances can be seen as one effect of the widening division in the population based on political creeds and affiliations. 

\subsection*{Non-isotropy of language embeddings}

Language isotropy has been thought of as a reason for the robustness of the {\tt word2vec} models and any embedding tool in general. Isotropy in a geometric sense is the measure of uniformity of the word embeddings across the inherent embedding space. The core idea that is assumed to support the word embeddings (and approaches based on them) is as follows: All natural languages must be able to describe all concepts in the language model using minimal combinations of words. This property is facilitated as the words become uniformly distributed across the space\cite{arora2015rand}.

Persistent homology offers an easy way to measure the isotropy of any word embedding model by looking at the point cloud of the embeddings. The presence of holes in the embedding space can be thought of as parts of the space which are poorly described using the current geometry and for which news words should either be introduced or words can be remapped to new meanings, reminiscent of Moran processes in evolution and linguistics \cite{tiefelsdorf2006modelling}.

We took the subreddit data from each of the three political subreddits and calculated the embeddings of the word corpus to get a representation of the words at the end of 2017. We observed the presence of multiple large homology groups suggesting inconsistencies with the hypothesis of isotropy of {\tt word2vec} embeddings. Our observation, albeit in a limited context provokes additional analysis of {\tt word2vec} models and their effectiveness. Another potential investigation is the location of the homologies and identifying the regions of space contributing to the homologies. This strategy may lead to a tool for analyzing a text corpus and identifying topics which can be misrepresented. Such a tool can point to potential pitfalls of the embeddings and also new approaches to avoid them.

\subsection*{Using user data to find similarities of subreddits}

One of the reasons for conducting the experiment on a per user basis is to be able to identify the communities from population data and minimal structural information. This new individualized data prompted us to re-perform the previous analysis of subreddit distance based on only the user data. We took the word corpus for each user and made an incremental {\tt word2vec} model to get temporal embeddings of the each user from June 2015 till November 2017. Using these embeddings, we calculated the average distance between each pairs of users in the subreddits to observe changes in the language representations.

The average user distance between the subreddits remained largely unchanged throughout the time period of analysis, painting a different picture than the more robust analysis from the overall subreddit data. This discrepancy prompts a more detailed analysis of using personalized data to gather succinct information to compare communities. This approach also faces a problem in identifying communities based on individualized data, where no proper means of learning the underlying population graph exists. In a setting where conversations take place with multiple users, the problem of inferring the communication hypergraph is a harder problem \cite{kim2017community}.

\subsection*{Intra-subreddit language drift using users}

\begin{figure}[ht]
    \centering
    \includegraphics[width=0.9\textwidth]{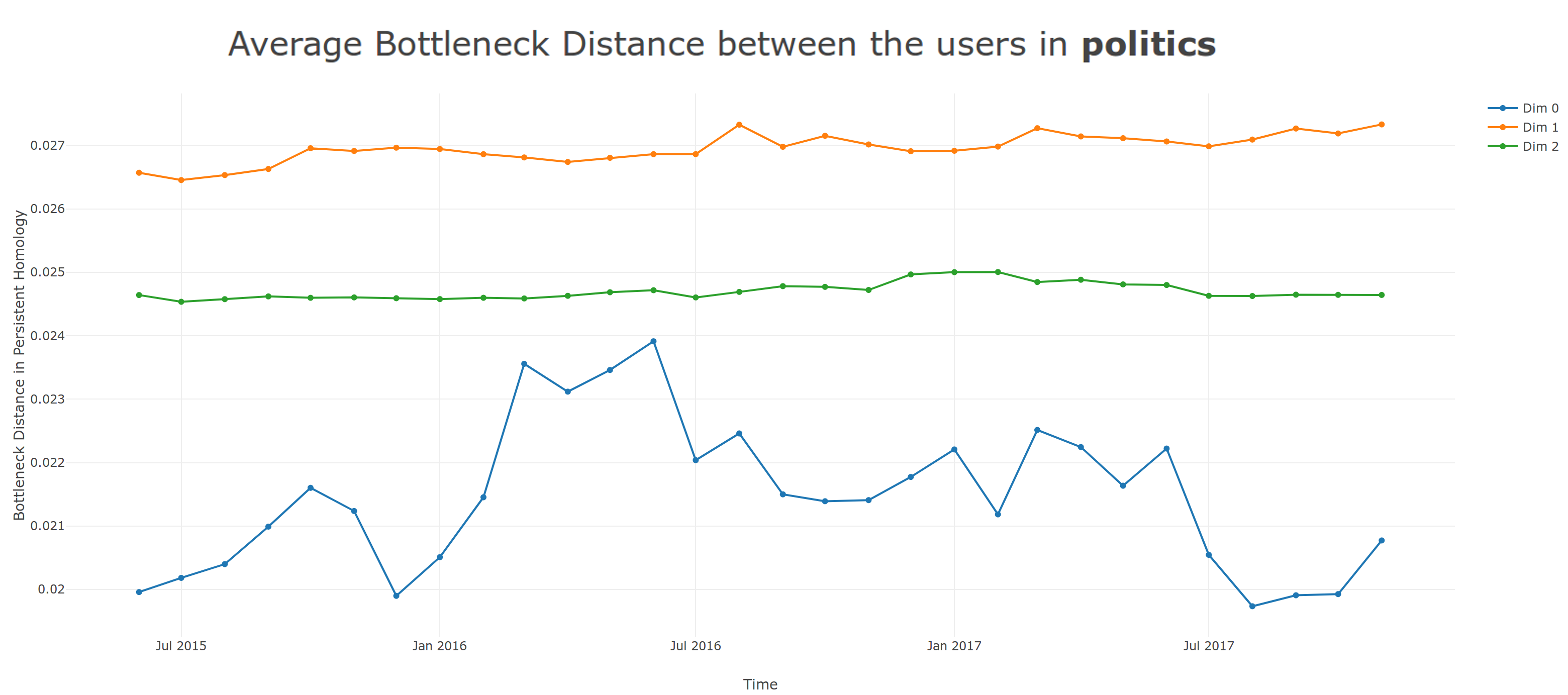}\\
    \includegraphics[width=0.48\textwidth]{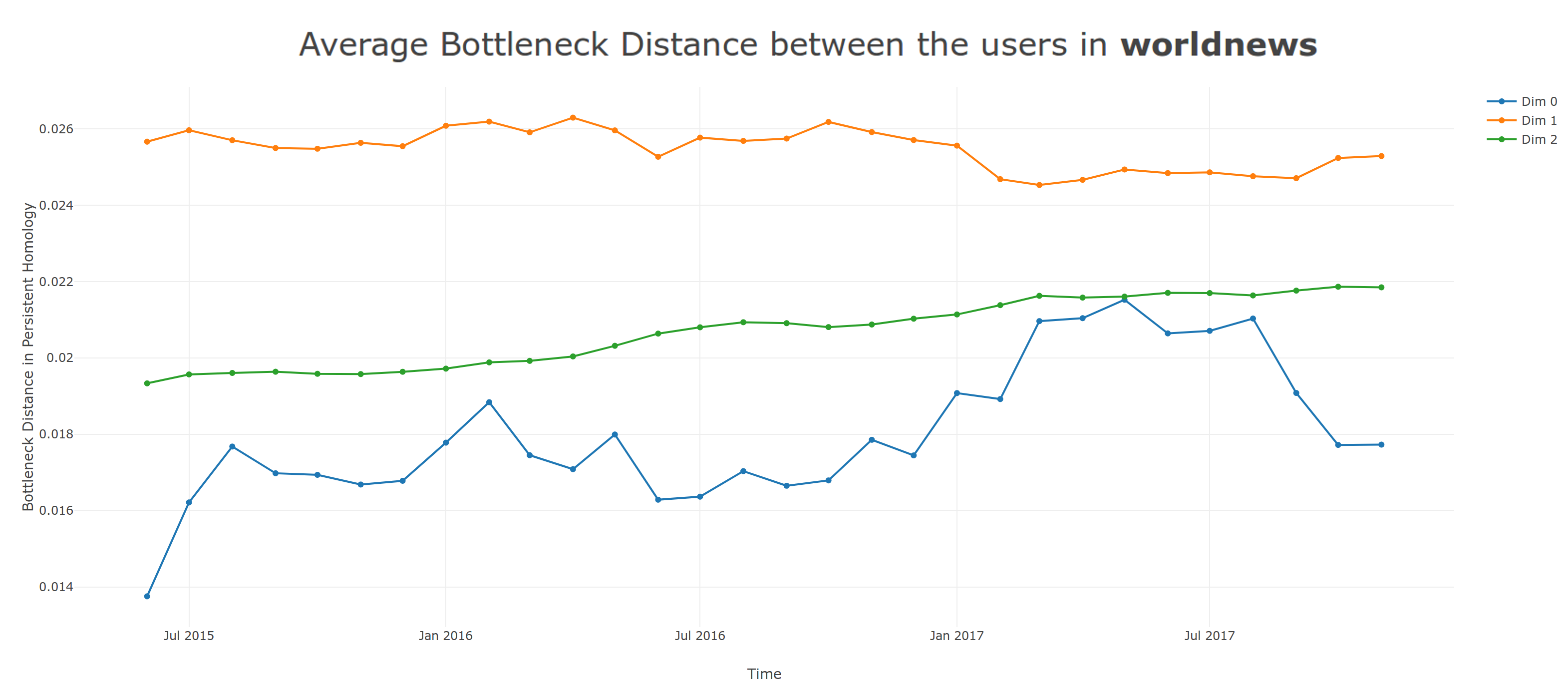}
    \includegraphics[width=0.48\textwidth]{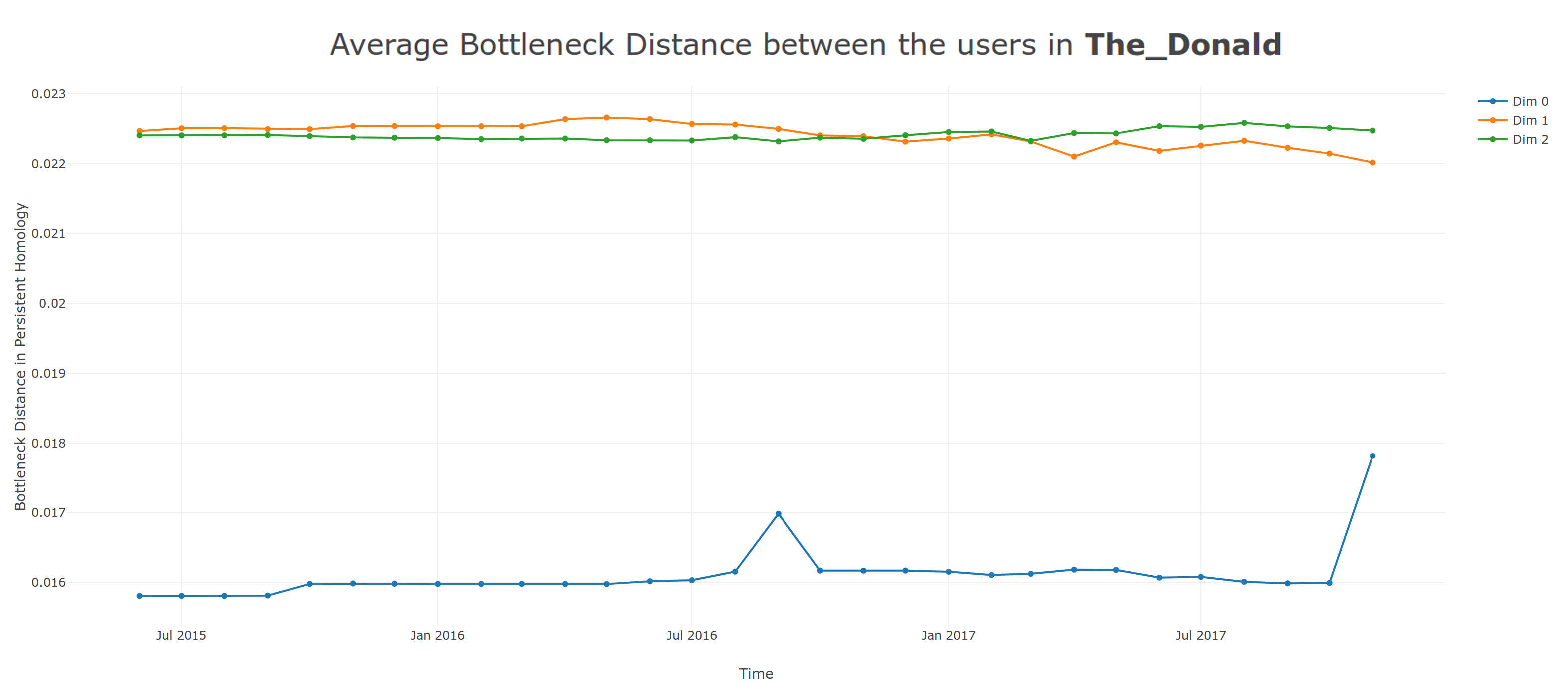}
    \caption{Intra-subreddit distance. The individualized embeddings inside each subreddit can help us understand the convergence of language over time and the stability of the language after convergence. The current user distances remain stable over a period of two years suggesting a stable distribution of language representations, where the divergence observed \emph{a priori} is an effect of the drift in language due to shifts in the topic focus over time.\label{fig:subdist2}}
\end{figure}

To observe the drift in language over time we examine the distance between the representations of each user over time (\cref{fig:subsdist}). The user data has many limitations, namely, initialization process is slow;  vocabulary remains limited; length of conversations is typically short; and most importantly, the best existing data corpus is inadequately small. Due to these limitations, any kind of user based analysis of subreddits has proven difficult. We notice a small pattern of increasing distance, reminiscent of the subreddit distance metric. But the fluctuations in first two homologies show the effect of lack of data on the bottleneck distance.

One way of getting around this limitation is to have robust user data to construct good individual representations of the language. The design of {\tt intuits} is such that the crowd-sourced natural experiments can yield better  individual representations, each of which can be tracked over time to get drift of the language and observe the community effects on the representation. Collecting more focused data, such as the ones to be gathered by the {\tt intuit} project,  will help reveal much more about various linguistic hypotheses -- ranging from origins of the language to its universality and stability. 

\section*{Discussion}

We conclude that design and launch of {\tt intuit}'s large-scale crowd-sourced creolization experiment constitutes a feasible project -- \emph{proviso}, serious attention is given to language's convergence properties (and subsequent stability). Our computational simulation of Bayesian Echo Chamber and the mathematical analysis of convergence to equilibria within it appear promising for the following reasons: (i) by providing the right tools to a crowd-sourced wiki-like public effort, it seems conceivable to creolize a natural language more suitable for the world-wide web and (ii) furthermore, by not ignoring the effects of naturally occurring population (graph) structures (e.g., reddit), it seems possible to avoid certain natural limitations, usually exhibited as disparate Echo Chambers, coexisting, but in fundamental disagreement with one another. Thus there must be significant efforts to bridge the differences between the idealized theoretical model and extant empirical models, which may be achieved by simply prompting conversations among key individuals, who could facilitate rapid mixing in the population graph. Theories of random graphs, expander graphs and algebraic analysis of graphs provide powerful mathematical tools to  achieve these goals algorithmically. 

We hypothesize further that a properly designed {\tt intuit} experiment will parametrize the universal grammar (assuming and validating its existence) common to natural languages; it will quickly converge to a highly stable Nash equilibrium; and it will optimize certain information-theoretic utility functions for the utterer-hearer pairs. These hypotheses are, separately and together, refutable. The data collected from this natural experiment will shed important light on the biological mechanisms responsible for the emergence of human languages, while spurring the emergence of a new wave of language creation.

The experiment also raises additional questions:

 {\em How will the intuit language relate to the ongoing research in Artificial Intelligence?} Currently there is much interest in using deep learning for natural language processing, especially for language translation, text-tagging, captioning images, etc. -- all relying on some form of {\tt word2vec} embeddings based on large corpora from multiple languages. There is a lack of a proper theory in deep learning explaining its spectacular successes and intriguing failures (e.g., adversarial perturbations) that this version of AI (sub-symbolic, black-boxes) exhibits. Our work on the signalling-game-theoretic models, as initiated here, could be useful in injecting robustness to the future AI research. A particularly colorful example of a confusing experiment in AI involves Microsoft's Tay, which was effortlessly hijacked by a millenials' echo chamber.
 
 {\em How will the intuit language relate to the current thinking in Mathematical Data Science?} We have shown here that topological analysis of point-cloud-data provides a powerful tool that could be widely applicable. Some applied works on evolutionary studies in virology and oncology have been influential, but wider applications remain unexplored, especially in the context of the evolution of languages, social norms, social contracts, social institutions, etc., -- all topics of immense importance as intelligence/information technologies have begun to disrupt long-standing, hitherto stable institutions in unpredictable manners. Creolization's deeper relations to topological data analysis (TDA), Manifold Learning, Information Geometry, Game Theory etc.~are thus important topics of future research.
 

{\em How will the intuit language relate to the current thinking in Biology?}
Our experiments anticipate support for the usefulness of  distributional methods of representing semantics in a language. Our approach is supported by the analysis by Arora et al.\cite{arora2015rand}, who were able to identify a semantically-relevant low-dimensional shared representation of fMRI responses. Their experiments and analysis were conducted in an unsupervised fashion and involved views of multiple subjects watching the same natural movie stimulus. These studies point to some fundamental questions about the biology of languages and how it evolved in a relatively short period. Our analysis using {\tt intuits} -- with its multimodal emoji like structures -- is hoped to raise more challenges and resolve ancient mysteries.

Last but not least, {\em how will the intuit language relate to the current thinking in Linguistics?} Noam Chomsky and his followers have played a dominant role in shaping the current theories of language, but in isolation from other evolutionary researchers and their theories, such as cellularization (codons), endosymbiosis, multi-cellularity, speciation,etc. However, human spoken language is hypothesized to be a biological artefact (postulating a yet-to-be identified language organ; related to the so-called I-language; and supporting distributional semantics), but leads to theories that are unexperimentable (``not-even-wrong''). The existence of WWW and crowd-sourcing drastically changes the situation by enabling scalable and experimental inventions of new artificial natural languages using large number of communicating human learners.

However, our biggest challenges will remain in the engineering of the {\tt intuit} Linguistic System, focusing on how the data should be collected and how it should be analyzed. We can use existing efforts developed in cloud computing (e.g., BigTable, BigQuery, etc.), enabling construction of such a system with relatively small man-power. But given that internet is already affecting how younger generations communicate (with hashtags, emojis, acronyms, etc.), the window of opportunity for the natural experiments based on {\tt intuit} may be closing soon, particularly as the field gets crowded by powerful monolithic corporations, namely, the so-called unicorns e.g. Twitter (tweets), Facebook (identity systems) and Google (Language Translations).
\section*{Methodology}

Here, we show the guarantees of the interactive model for language evolution. We apply the model to real world data and show how the properties of the model give us insights into the data using persistent homology. For more details, see the supplementary materials.

\subsection*{Modeling Change in Language}

\begin{table*}
\centering
\small
\caption{Translation between persistent homology and linguistic terminology\label{tab:trans}}
\begin{tabular}{ll}
Persistent homology & Linguistic interpretation\\
\midrule
Filtration value $\epsilon$ & Clustering of words up to similarity of $\epsilon$\\
Persistence diagrams & Representation of difference in similarities of a \\
& word to its semantic neighbors\\
$0$-dimensional betti numbers at distance $\epsilon$ & Number of word clusters based on semantic\\
& dis-similarity up to $\epsilon$\\
Generators of $0$-dimensional persistence diagram & Representatives of word clusters\\
Birth-death timings of $0$-dimensional generators & Hierarchical clustering of words\\
$1$-dimensional persistence diagram & Small cycles help in detection of polysemous words\\
Distance between persistence diagrams & Measure of difference of two word corpus, in terms\\
& of semantic meanings associated to words\\
Higher dimensional persistence diagrams & Non-isotropy of word embeddings\\
Generators of higher dimensional & Bounding regions of space with no embeddings\\
persistence diagrams & contradicting isotropy\\
\bottomrule
\end{tabular}
\end{table*}

Many different representation scheme can be used to model a language \cite{mnih2007three}\cite{collobert2008unified}\cite{white2003nature}. We represent vocabulary of a language in a high-dimensional space, with the properties that (a) Similar words/intuits (based on their contextual usage or annotations) must be placed in nearby region in this space; (b) Any change due to evolution of language can be captured through the relative movements of the words in this space.

\subsection*{Bayesian Echo Chamber}

This is a  new \emph{Bayesian generative model} for social interaction data, for uncovering influence-relations among individuals from their time-stamped conversation data\cite{guo2015bayesian}. The forcing function of a social process is based on the mutual influence among the participants, which must be inferred from the flow of the interaction \cite{reali2009words}. Evolution in a language can be modeled to explain temporal stability of the word meanings, despite their  occasional misuses -- but also, sporadic changes in meanings that do occur within subcommunites (Echo Chambers) and propagate further\cite{pelikan2000bayesian}. They may be assumed to be accompanied by concomitant changes in usage and grammatical structure to connect these words.Social influence forces these changes and the flow of changes originates from the ``most influencing'' participant to the ``weaker participant''. We use Bayesian Echo Chamber to understand how external influences, mixing of different social and linguistic cultures in a speech community initiate language evolution and how it finally converges to a stabilized form, where after stability it is less susceptible to external stimulants.

\subsection*{Persistent Homology}

We base our analysis on the field of algebraic topology, which can capture the global differences between two high dimensional embeddings as well as give local information about a representation to get insights into distribution of the words (or intuits) in the space. We measure ``features'' of the space that remain invariant under continuous deformations, such as stretching, bending, rotating but not tearing or gluing parts of the space. These features correspond to the ``holes'' in the space, that range across all dimensions of the space and can capture higher order structures -- more so, than simple combinations of elementary structures, which are commonly used in machine learning.\\
We create a continuous topological space from a corpus of points, by putting balls of size $\epsilon$ around each point. The union of these balls gives the \v{C}ech complex, which has the exact topology of the underlying space but is hard to compute. We instead focus on the Vietoris-Rips(VR) complex, which is a smaller simplicial complex but can be shown to be a good approximation of the \v{C}ech complex\cite{carlsson2009topology}\cite{edelsbrunner2014short}. The VR complex of size $\epsilon$ for a corpus $C$, $VR_\epsilon(C)$, is constructed by joining two points if they have a distance smaller than $\epsilon$, which corresponds to the threshold of word similarities. By building the $VR_\epsilon(C)$ complex at different scales of $\epsilon$, we can see when the holes are generated and when they are filled up, called the \emph{birth} and \emph{death} ``times'' of the holes (homologies). These are encoded in the barcode representation \cref{fig:tdaoutline} and also as points on the plane, known as a persistence diagram.\\
The presence of holes across different dimensions gives us an estimate of how different two spaces are. Topologically, we can think of this as the obstructions from changing one space into another using continuous methods. Persistent homology strengthens these notions and enables us to capture more information about these holes, such as the sizes and the boundary words of the holes. The longer a barcode is, the higher the probability that the underlying feature is a characteristic of the manifold and not a by-product of noise in the data\cite{edelsbrunner2000topological}\cite{guskov2001topological}. We use a topological metric called \emph{the bottleneck distance} between persistence diagrams, which is a special case of the Wasserstein metric on this space. The bottleneck distance can be thought of as a correspondence between the homologies to minimize the disparity between spaces. Hence a large bottleneck distance is a proof of dissimilarity between the underlying spaces. Moreover, the generators of the corresponding homology groups helps identify the local structures generating the deformations\cite{chazal2015convergence}. Overall, we connect the ideas from algebraic topology, language evolution and computational linguistics, and provide a map to decipher notions interchangeably between these fields in \cref{tab:trans}.

\ACKNOWLEDGMENT{We would like to thank Sylvain Cappel, Misha Gromov of Courant Institute of Mathematical Sciences, Raul Rabadan of Columbia University, Rohit Parikh of CUNY, Larry Rudolph of TwoSigma for their keen insight and enlightening discussions in the makings of this project. We would also like to thank Halley Young of CMU who kickstarted our project in its infancy and get better understandings of the datasets available.\\
B.M. was supported by an Army Research Office Grant \#A18-0613-001.
}%

\bibliographystyle{informs2014}
\bibliography{biblio.bib}

\end{document}